\documentclass[sn-mathphys]{sn-jnl}% Math and Physical Sciences Reference Style
\usepackage{indentfirst}
\usepackage{graphicx}
\usepackage{subfigure}
\usepackage{float}
\usepackage{etoolbox}
\usepackage{bbding}
\usepackage{pifont}
\usepackage{chngpage}
\usepackage{makecell}

\usepackage{array}
\usepackage{caption}
\makeatletter
\patchcmd{\ps@headings}
{\hbox to \hsize{\hfill Springer Nature 2021 \LaTeX\ template\hfill}}
{\hbox to \hsize{}}
{}
{}
\patchcmd{\ps@titlepage}
{\hbox to \hsize{\hfill Springer Nature 2021 \LaTeX\ template\hfill}}
{\hbox to \hsize{}}
{}
{}
\makeatother
\makeatletter
\patchcmd{\ps@headings}
{\hbox to \hsize{\hfill Springer Nature 2021 \LaTeX\ template\hfill}}
{\hbox to \hsize{}}
{}
{}
\patchcmd{\ps@headings}
{\hbox to \hsize{\hfill Springer Nature 2021 \LaTeX\ template\hfill}}
{\hbox to \hsize{}}
{}
{}
\patchcmd{\ps@titlepage}
{\hbox to \hsize{\hfill Springer Nature 2021 \LaTeX\ template\hfill}}
{\hbox to \hsize{}}
{}
{}
\makeatother

\begin{document}	

\title[Article Title]{A Multi-scale Fused Graph Neural Network with Inter-view Contrastive Learning for Spatial Transcriptomics Data Clustering\\
}

\author{
	% ??1
	\begin{minipage}[t]{0.28\textwidth} % ?????????????
		\centering
		\textbf{1\textsuperscript{st} Jianping Mei}\\[2pt]
		\footnotesize
		\textit{College of Computer and}\\
		\textit{Information Science}\\
		\textit{Southwest University}\\
		Chongqing, China\\
		\texttt{mjp15207948724@163.com}
	\end{minipage}
	% ??2
	\begin{minipage}[t]{0.28\textwidth}
		\centering
		\textbf{2\textsuperscript{nd} Siqi Ai}\\[2pt]
		\footnotesize
		\textit{College of Computer and}\\
		\textit{Information Science}\\
		\textit{Southwest University}\\
		Chongqing, China\\
		\texttt{599900503@qq.com}
	\end{minipage}
	% ??3 (????)
	\begin{minipage}[t]{0.28\textwidth}
		\centering
		\textbf{3\textsuperscript{rd} Ye Yuan}\textsuperscript{*}\\[2pt]
		\footnotesize
		\textit{College of Computer and}\\
		\textit{Information Science}\\
		\textit{Southwest University}\\
		Chongqing, China\\
		\texttt{yuanyekl@swu.edu.cn}\\[4pt]
		\tiny *Corresponding author
	\end{minipage}
}

%\author[1,2]{\fnm{} \sur{Jianping Mei}}
%\author[1,3]{\fnm{} \sur{Xiangli Li}}\email{lixiangli@guet.edu.cn}
%\author[1]{\fnm{} \sur{Yuanjian Mo}}

%\affil[1]{\orgdiv{School of Mathematics and Computing Science}, \orgname{Guilin University of Electronic Technology}, \city{Guangxi}, \postcode{541004}}
%\affil[2]{\orgdiv{Guangxi Colleges and University Key Laboratory of Data Analysis and Computation}, \city{Guangxi}, \postcode{541004}}
%\affil[3]{\orgdiv{Center for Applied Mathematics of Guangxi}, \city{Guangxi}, \postcode{541004}}

\abstract{Spatial transcriptomics enables genome-wide expression analysis within native tissue context, yet identifying spatial domains remains challenging due to complex gene-spatial interactions. Existing methods typically process spatial and feature views separately, fusing only at output level - an "encode-separately, fuse-late" paradigm that limits multi-scale semantic capture and cross-view interaction. Accordingly, stMFG is proposed, a multi-scale interactive fusion graph network that introduces layer-wise cross-view attention to dynamically integrate spatial and gene features after each convolution. The model combines cross-view contrastive learning with spatial constraints to enhance discriminability while maintaining spatial continuity. On DLPFC and breast cancer datasets, stMFG outperforms state-of-the-art methods, achieving up to 14\% ARI improvement on certain slices.}


\maketitle
\section{Introduction}\label{sec1}

Spatial Transcriptomics (ST) technologies have revolutionized the study of tissue architecture, creating a paradigm shift that allows researchers to interrogate genome-wide expression information directly within its original morphological context \cite{b1}. Identifying spatial domains defined by coherent gene expression in contiguous tissue regions is essential for understanding development, disease progression, and cellular interactions. While clustering methods have been widely adopted for this task, early approaches relying on conventional algorithms or shallow spatial constraints fail to capture complex biological relationships \cite{b2,b3}.

\begin{figure*}[htbp]
	\centering
	\includegraphics[width=11.8cm]{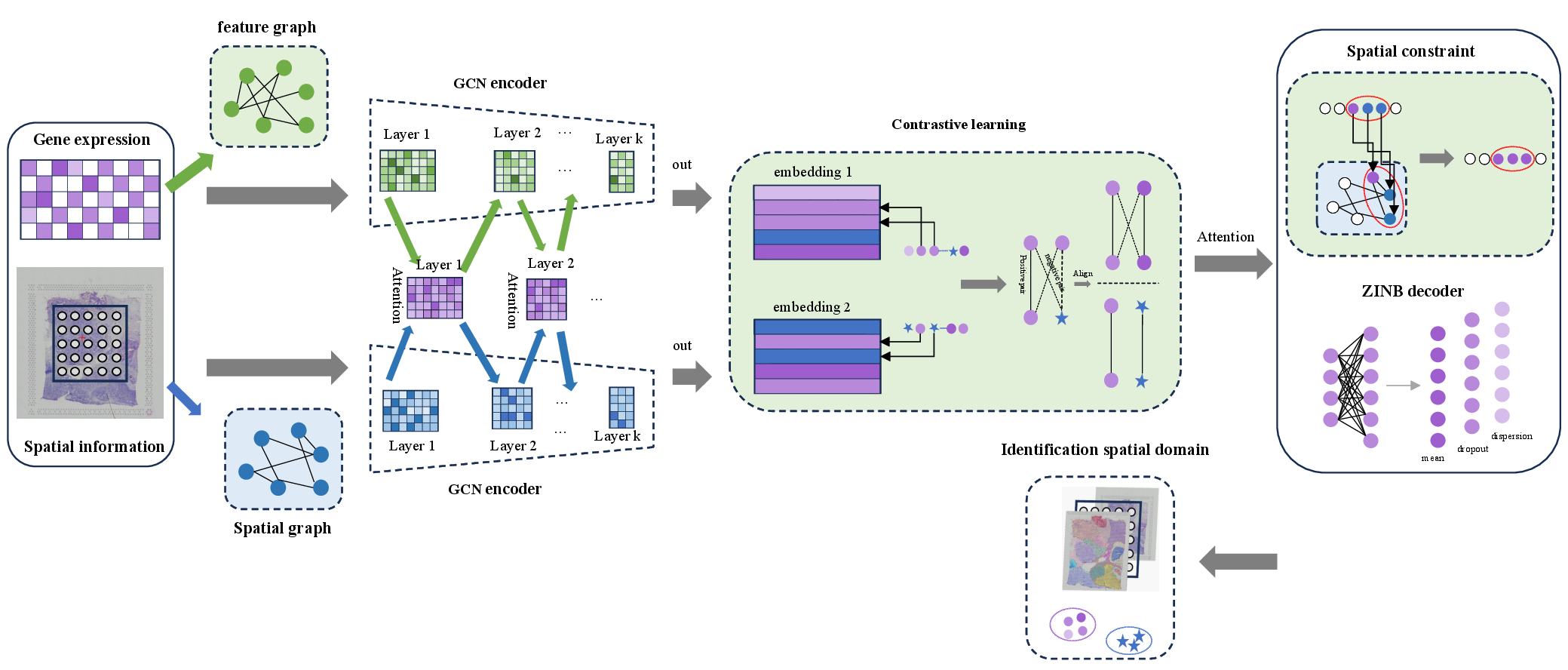}
	%\caption{fig1}
	\caption{ The framework of stMFG. By learning the multi-scale representations of Spaces and feature views through different GCN encoders, the attention mechanism is utilized to fuse each layer of embeddings between views, and the unified fusion is used as the input for the next layer of GCN of different views. Contrastive learning between views and spatial constraints achieve high-level discriminative feature learning and semantic alignment while maintaining organizational spatial continuity. The original data is reconstructed using the ZINB decoder, and the final complementary discriminant embedding is used for spatial domain recognition.}
	\label{fig:1}
\end{figure*}

Recent graph neural network (GNN) \cite{b4,b5,b6,b7,b8,b9} methods model spatial neighborhoods and gene expression through dual-view graphs. However, most methods process these views separately and fuse embeddings only at the output layer, limiting multi-scale feature interaction and representation power \cite{b10,b11,b12}. To overcome these limitations, stMFG is proposed, a multi-scale attention-fusion graph neural network. Our key innovation is a layer-wise interactive fusion mechanism that dynamically integrates spatial and feature information after each graph convolution. Additionally, stMFG incorporates cross-view contrastive learning to enhance representation consistency, spatial regularization to maintain domain continuity, and a zero-inflated negative binomial (ZINB) decoder to model gene expression characteristics. Comprehensive experiments show stMFG outperforms state-of-the-art methods in clustering accuracy and spatial domain identification, demonstrating its effectiveness for ST data analysis.The following are the contributions of stMFG:
\begin{itemize}
	\item A cross-view multi-scale deep fusion module is designed to dynamically integrate spatial and feature information after each graph convolutional layer, overcoming the limitation of traditional methods that only fuse information at the output layer, thereby enabling deep interaction of multi-scale semantic features.
	\item By seamlessly integrating cross-view contrastive learning with spatial regularization constraints, the model enhances representational discriminability while preserving tissue spatial continuity, forming a multi-task collaborative end-to-end learning paradigm.
	\item Experimental results on multiple benchmark datasets validate the superior performance of stMFG.
\end{itemize}

\section{Methodology}
\subsection{Overview}\label{AA}
The stMFG framework identifies spatial domains in transcriptomics data through a novel architecture. Its core is a multi-scale interactive fusion mechanism that uses layer-wise cross-view attention to dynamically integrate spatial and gene expression information. Supplemented by contrastive learning and biological constraints, the model captures both local and global tissue structures with high biological plausibility, as shown by Fig. \ref{fig:1}.

\subsection{Graph Construction}\label{AA}
To comprehensively capture the heterogeneous characteristics of spatial transcriptomics data, we construct two distinct graph structures that encode complementary aspects of cellular organization \cite{b11,b12}. 

The spatial graph $G_s = (A_s, X)$ encapsulates tissue architecture, with adjacency matrix $A_s \in \mathbb{R}^{N \times N}$ constructed based on Euclidean distance \cite{b11}:
\begin{equation}
A_s^{ij} = 
\begin{cases} 
1, & \text{if } i \in \mathcal{N}_j \ \text{or} \ j \in \mathcal{N}_i \\
0, & \text{otherwise}
\end{cases}
\end{equation}
where $\mathcal{N}_i$ denotes the spatial neighbors of spot $i$, determined by a predefined radius $r = 550$.

The feature graph $G_f = (A_f, X)$ captures gene expression similarity via KNN based on cosine similarity \cite{b11}.

\subsection{Multi-scale Deep Interactive Fusion}\label{AA}
A multi-scale interactive fusion mechanism is proposed for hierarchical integration of spatial and feature information across network layers.

Let $Z^l \in \mathbb{R}^{N \times d_l}$ denote the fused embedding at layer $l$, with $Z^0 = X$. For each layer $l = 0, 1, \ldots, L-1$:

\noindent\textbf{Dual-view GCN}: Spatial and feature GCN encoders generate view-specific representations \cite{b12}:
\begin{equation}
Z_s^l = \sigma\left(\tilde{D}_s^{-\frac{1}{2}}\tilde{A}_s\tilde{D}_s^{-\frac{1}{2}} Z^l W_s^l \right),
\end{equation}
\begin{equation}
Z_f^l = \sigma\left(\tilde{D}_f^{-\frac{1}{2}}\tilde{A}_f\tilde{D}_f^{-\frac{1}{2}} Z^l W_f^l \right),
\end{equation}
where $\tilde{A} = A + I$, $\tilde{D}$ is the degree matrix, $W^l$ are learnable parameters, and $\sigma$ is ReLU.

\noindent\textbf{Cross-view Attention Fusion}: Dynamic integration of view-specific embeddings is achieved through \cite{b13}:
\begin{equation}
M^l = \ell_2\left(\text{softmax}\left(\text{LeakyReLU}\left([Z_s^l \| Z_f^l] W_a^l \right)\right)\right),
\end{equation}
where $W_a^l \in \mathbb{R}^{2d_l \times 2}$, and $M^l = [m_s^l \| m_f^l] \in \mathbb{R}^{N \times 2}$.

The fused embedding for next layer:
\begin{equation}
Z^{l+1} = (m_s^l \mathbf{1}_{d_l}) \odot Z_s^l + (m_f^l \mathbf{1}_{d_l}) \odot Z_f^l,
\end{equation}
where $\mathbf{1}_{d_l} \in \mathbb{R}^{1 \times d_l}$ and $\odot$ denotes Hadamard product.

\noindent{Final unified embedding after $L$ layers is denoted as $Z = Z^L$.}

\subsection{Cross-view Contrastive Learning}\label{AA}
Cross-view contrastive learning \cite{b14} is employed to enhance representation discriminability and view consistency.

For each spot $i$, paired embeddings $(z_s^i, z_f^i)$ are treated as positive samples. Negative pairs include cross-view $(z_s^i, z_f^j)$ and within-view $(z_s^i, z_s^j)$, $(z_f^i, z_f^j)$ for $j \neq i$. The contrastive loss \cite{b14} :
\begin{equation}
\begin{aligned}
\mathcal{L}_{cl} = -\frac{1}{2N}\sum_{i=1}^N \Bigg[ 
& \log \frac{e^{(\text{sim}(z_s^i, z_f^i)/\tau)}}{\sum\limits_{k=1}^{N}\sum\limits_{v \in \{s,f\}} e^{(\text{sim}(z_s^i, z_v^k)/\tau)} - e^{(1/\tau)}} \\
& + \log \frac{e^{(\text{sim}(z_f^i, z_s^i)/\tau)}}{\sum\limits_{k=1}^{N}\sum\limits_{v \in \{s,f\}} e^{(\text{sim}(z_f^i, z_v^k)/\tau)} - e^{(1/\tau)}} \Bigg],
\end{aligned}
\end{equation}
where $\text{sim}(\cdot,\cdot)$ is cosine similarity, $\tau$ denotes temperature parameter.

\subsection{Spatial Regularization}\label{AA}
Spatial regularization is imposed to maintain neighborhood relationships in latent space \cite{b12}:
\begin{equation}
\mathcal{L}_{reg} = -\sum_{i=1}^N \left( \sum_{j \in \mathcal{N}_i} \log \sigma(C_{ij}) + \sum_{k \notin \mathcal{N}_i} \log (1 - \sigma(C_{ik})) \right),
\end{equation}
where $C$ is cosine similarity matrix from $Z$, $\mathcal{N}_i$ denotes spatial neighbors of spot $i$, and $\sigma(\cdot)$ is sigmoid function.

\subsection{ZINB-based Reconstruction}\label{AA}
A ZINB decoder \cite{b15} is used to model zero-inflation and over-dispersion in gene expression data \cite{b15}:
\begin{equation}
\text{ZINB}(x \mid \pi, \mu, \theta) = \pi \cdot \delta_0(x) + (1-\pi) \cdot \text{NB}(x \mid \mu, \theta).
\end{equation}
Reconstruction loss:
\begin{equation}
\mathcal{L}_{zinb} = -\frac{1}{N \times M}\sum_{i=1}^N \sum_{j=1}^M \log \text{ZINB}(x_{ij} \mid \pi_{ij}, \mu_{ij}, \theta_{ij}),
\end{equation}
where parameters are estimated from $Z$ by the decoder. Given that ST data are non-negative counts, thus early methods  extensively use non-negative matrix factorization and its variants \cite{b16,b17,b18,b19,b20,b21,b22,b23,b24,b25,b26,b27,b28,b29,b30,b31,b32,b33,b34,b35,b36,b37,b38,b39,b40,b41,b42,b43,b44,b45,b46,b47,b48,b49,b50,b51,b52,b53,b54} to extract low-dimensional features.

\subsection{Model Training }\label{AA}
The complete loss functione is defined as:
\begin{equation}
\mathcal{L} = \alpha \mathcal{L}_{zinb} + \lambda \mathcal{L}_{cl} + \gamma \mathcal{L}_{reg},
\end{equation}
where $\alpha$, $\lambda$, and $\gamma$ are hyperparameters that balance the contributions of reconstruction, contrastive learning, and spatial constraint.

\section{Experiments}

\begin{table}	
	\renewcommand{\arraystretch}{0.8}
	\centering
	\caption{The statistics of the datasets}\label{tab:1}
	\begin{scriptsize}
		\resizebox{0.45\textwidth}{!}{		
			\begin{tabular}{ccccccccc}
				\toprule						
				Datasets &Spots &Genes& Domains\\
				\toprule 
				$151510$ & 4595
				& 33538 & 7   \\  
				$151669$ & 3636 & 33538 & 5   \\ 
				$151670$ & 3484 & 33538 & 5   \\ 
				$151671$ & 4093 & 33538 & 5   \\ 
				$151672$ & 3888 & 33538 & 5   \\ 
				HBC & 3798 & 36601 & 20   \\ 
				\bottomrule	
		\end{tabular}}
	\end{scriptsize}
\end{table}

\subsection{Datasets and Baselines}\label{AA}
\noindent \textbf{Datases:} The datasets used in the experiment included DLPFC \cite{b11} and Human Breast Cancer(HBC) \cite{b12}, all from the 10x Visium platform. 

Data were preprocessed using the SCANPY toolkit \cite{b55} to remove off-tissue spots and filter low-expression or low-variance genes, with dataset details provided in Table \ref{tab:1}.

\noindent \textbf{Baselines:} The three benchmarks are Spatial-MGCN \cite{b11}, MAFN \cite{b12} and SCANPY \cite{b55} respectively. 

\subsection{Experimental Setup}
The hyperparameters $\alpha$, $\lambda$ and $\gamma$ for stMFG were tuned from \{0.001, 0.01, 0.1, 1, 10\}. Training used a learning rate \cite{b56,b57,b58,b59,b60,b61,b62,b63,b64,b65,b66,b67,b68,b69,b70,b71} of 0.001, weight decay of 5e-4, and 200 epochs, with benchmarks run under their default settings. Cluster quality was evaluated using the ARI \cite{b72} and NMI \cite{b73,b74,b75,b76,b77,b78,b79}, with higher scores being preferable. The experiments ran on an AMD Ryzen 9 7945HX/NVIDIA RTX 4060 platform using PyTorch 2.4.0.

\subsection{Performance}
\begin{figure*}[htbp]
	\centering
	\subfigure[ ]{
		\includegraphics[width=1.2cm]{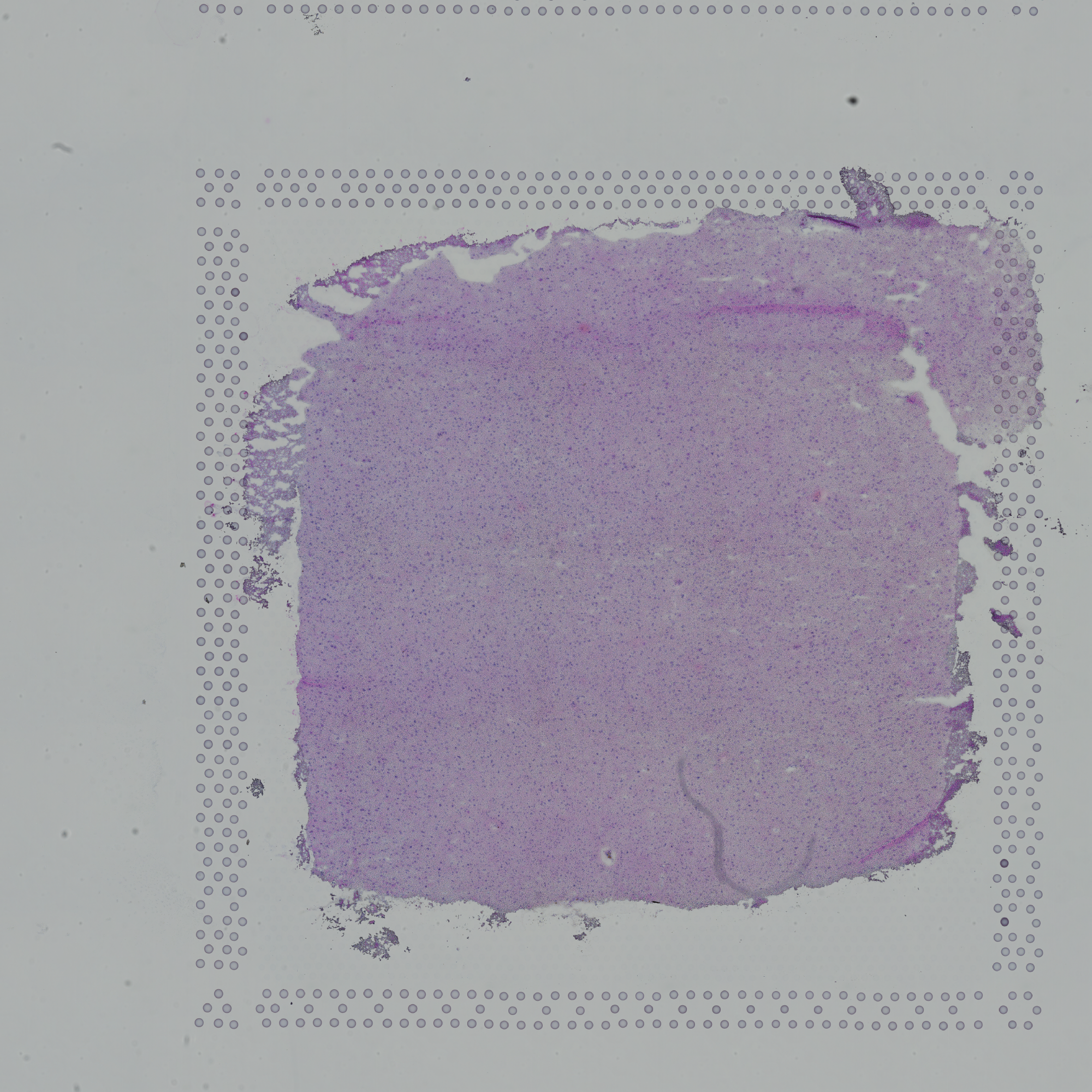}
		%\caption{fig1}
	}
	\quad
	\subfigure[ ]{
		\includegraphics[width=1.75cm]{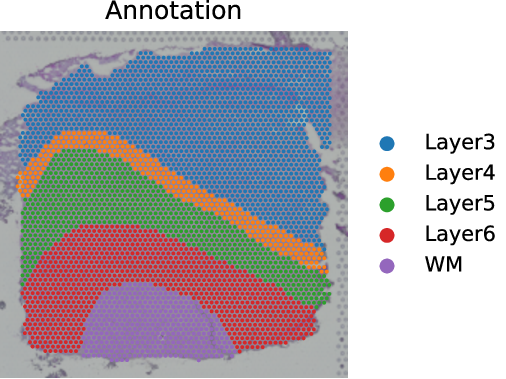}
	}
	\quad
	\subfigure[ ]{
		\includegraphics[width=1.2cm]{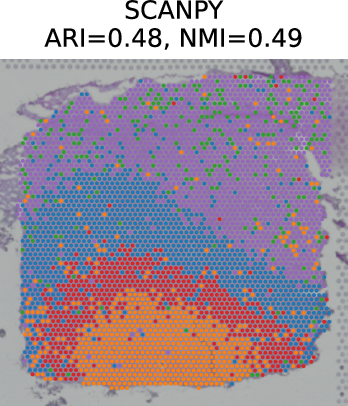}
	}
	\quad
	\subfigure[ ]{
		\includegraphics[width=1.2cm]{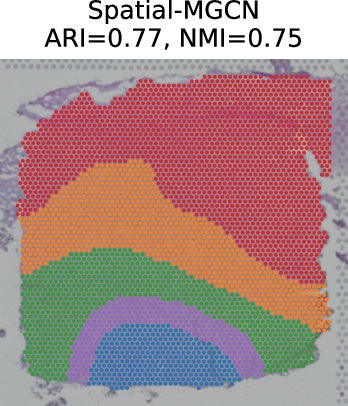}
	}
	\quad
	\subfigure[ ]{
		\includegraphics[width=1.2cm]{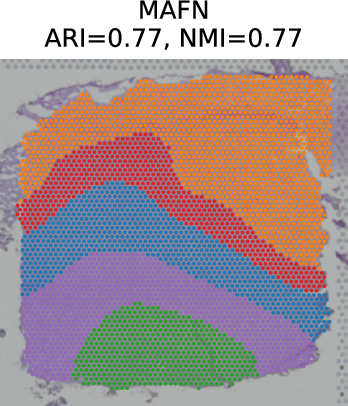}
	}
	\quad
	\subfigure[ ]{
		\includegraphics[width=1.2cm]{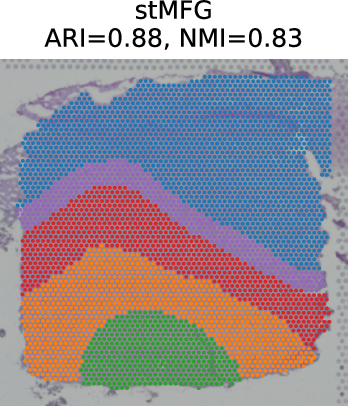}
	}
	\caption{ stMFG identifies spatial domains on the DLPFC dataset. (a) H$\&$S image for the 151672 slice, (b) Manual annotation. (c)-(f) Spatial domains are detected by SCANPY, Spatial-MGCN, MAFN and stMFG in 151672. }\label{fig:2}
\end{figure*}

Comprehensive evaluation on the DLPFC and HBC datasets demonstrates stMFG's superior clustering performance (Table \ref{tab:2}). On DLPFC slices with clear layered structures (e.g., 151671 and 151672), stMFG achieved outstanding results (ARI: 0.89, 0.88; NMI: 0.82, 0.83), surpassing the second-best method MAFN by approximately 10\%-14\% in ARI and 6\%-8\% in NMI. This performance is partly attributable to their simpler annotation scheme (5 regions per slice). stMFG also maintained robust results on more complex slices (151510) and the HBC dataset. Visualization in Fig. \ref{fig:2} confirms that stMFG generates spatial domains with clear boundaries and excellent continuity, precisely capturing subtle structural variations, especially in complex transitional areas.

\begin{table*}	
	\renewcommand{\arraystretch}{0.8}
	\centering
	\caption{Spatial clustering performance of different methods on DLPFC and human breast cancer}  \label{tab:2}
	\begin{scriptsize}
		\resizebox{0.80\textwidth}{!}{		
			\begin{tabular}{ccccccccc}
				\toprule						
				\thead{Evaluation\\ metrics}&Algorithm &$151510$& $151669$&$151670$&$151671$&$151672$&HBC&\\
				\toprule%\midrule 
				&  SCANPY & 0.27 & 0.26 & 0.24 & 0.41 & 0.48 & 0.60 \\  
				ARI&  Spatial-MGCN & 0.53 & 0.35 & 0.32 & 0.60 & 0.77 & 0.64 \\ 
				&  MAFN & 0.61 & \textbf{0.55} & 0.48 & 0.81 & 0.77 & 0.61 \\ 
				&  stMFG & \textbf{0.65} & 0.53 & \textbf{0.49} & \textbf{0.89} & \textbf{0.88} & \textbf{0.65} \\ 
				\toprule%\midrule
				&  SCANPY & 0.36 & 0.39 & 0.33 & 0.46 & 0.49 & 0.60 \\ 
				NMI&  Spatial-MGCN & 0.67 & 0.58 & 0.55 & 0.73 & 0.75 & \textbf{0.68} \\ 
				&  MAFN & 0.68 & 0.61 & 0.60 & 0.77 & 0.77 & 0.67 \\ 
				&  stMFG & \textbf{0.68} & \textbf{0.64} & \textbf{0.62} & \textbf{0.82} & \textbf{0.83} & \textbf{0.68} \\ 
				\bottomrule	
		\end{tabular}}
	\end{scriptsize}
\end{table*}

\begin{table*}	
	\renewcommand{\arraystretch}{0.8}
	\centering
	\caption{Ablation experiments to measure the influence of different components of stMFG on 151672}  \label{tab:3}
	\begin{scriptsize}
		\resizebox{0.7\textwidth}{!}{		
			\begin{tabular}{ccccccccc}
				\toprule						
				Performance & w/o mf
				& w/o $\mathcal{L}_{\rm{cl}}$& 
				w/o $\mathcal{L}_{\rm{reg}}$ & w/o $\mathcal{L}_{\rm{zinb}}$ & atMFG
				\\
				\toprule 
				ARI & 0.78
				& 0.81 & 0.85 & 0.79 & \textbf{0.88}  \\  
				NMI & 0.76 & 0.78 & 0.80 & 0.78 & \textbf{0.83}  \\ 
				\bottomrule	
		\end{tabular}}
	\end{scriptsize}
\end{table*}

\begin{figure}[htbp]
	\centering
	\subfigure[]{
		\label{fig:subfig_2a}
		\includegraphics[width=3.5cm]{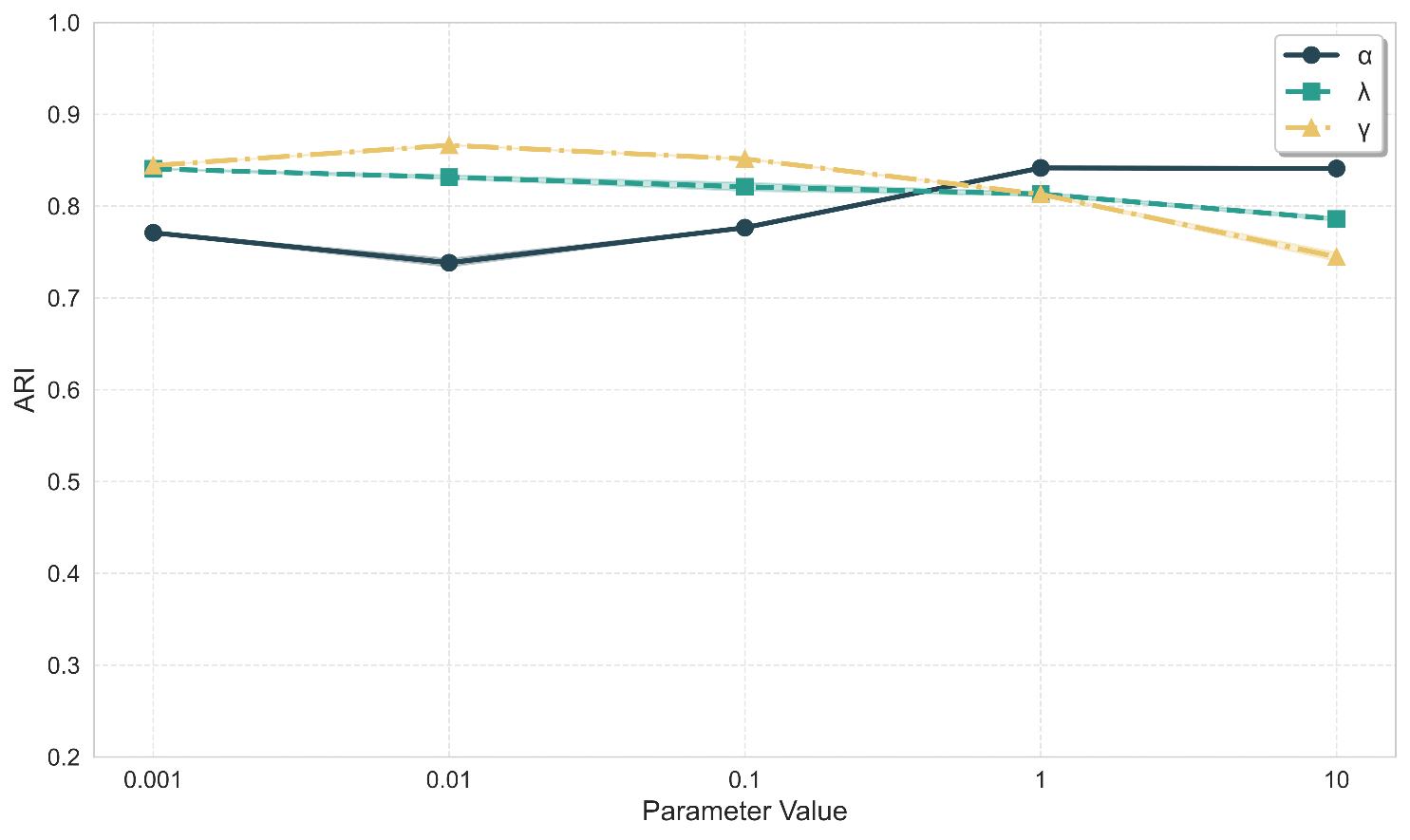}
		%\caption{fig1}
	}
	\quad
	\subfigure[]{
		\label{fig:subfig_2b}
		\includegraphics[width=3.5cm]{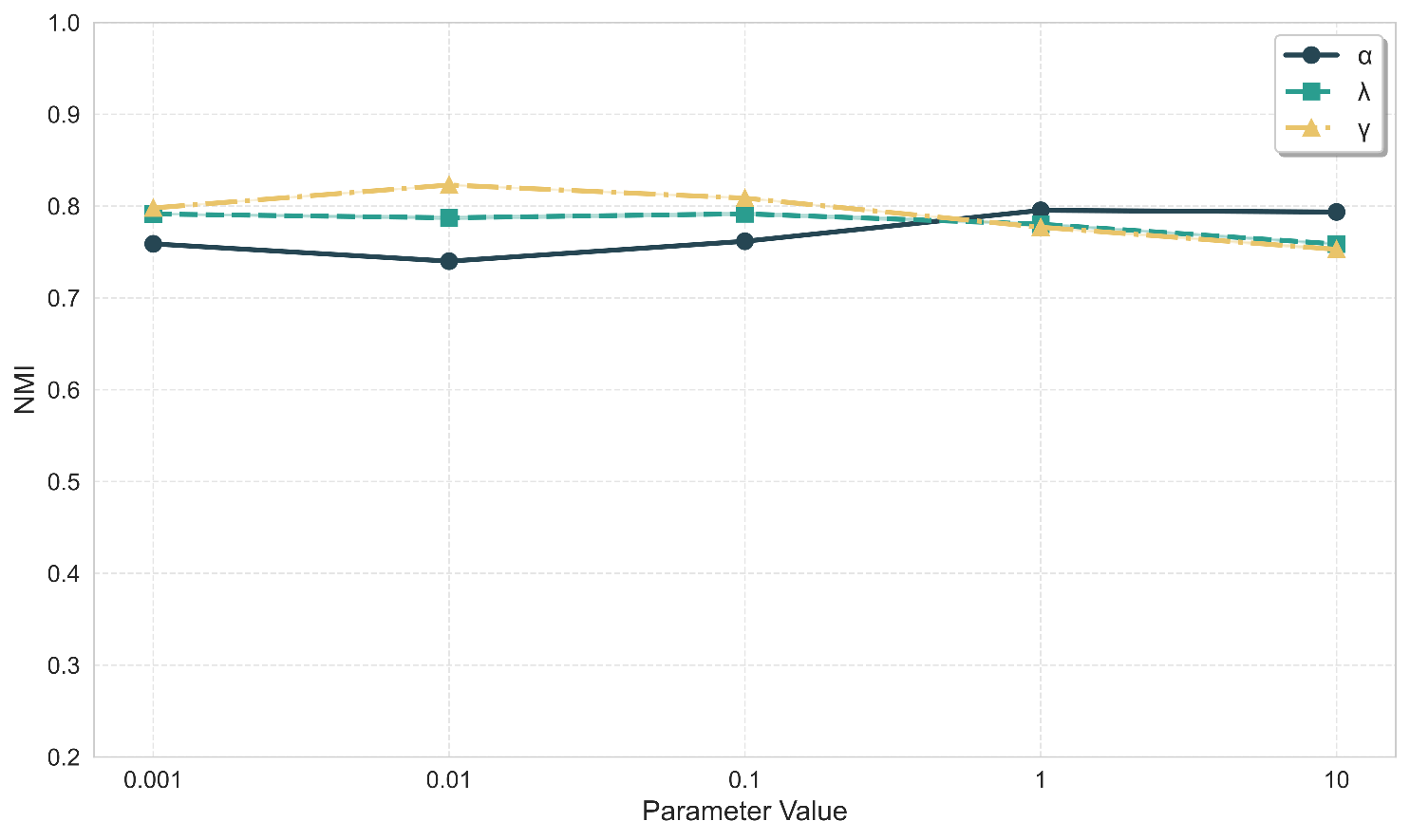}
	}
	\caption{ The results of hyperparameter sensitivity analysis on 151672}
	\label{fig:3}
\end{figure}

\subsection{Analysis of stMFG}
\noindent \textbf{Hyperparameter:} A parameter sensitivity analysis was conducted on slice 151672. According to the results shown in Fig. \ref{fig:3}, the model was observed to achieve optimal performance when $\alpha, \gamma$ was set to 1, 0.01 respectively. Meanwhile, $\lambda$ was found to be insensitive to specific values but was considered unsuitable for excessively large settings, it was set to 0.001.

\noindent \textbf{Ablation experiment:} An ablation study was conducted by systematically removing individual components: the multi-scale fusion (replaced with late fusion, ``w/o mf"), contrastive learning loss, spatial regularization, and ZINB loss. Results in Table \ref{tab:3} demonstrate that the absence of any component causes a significant performance drop, confirming that each contributes critically to stMFG's overall effectiveness.

\section{Conclusion}
In this paper, we propose stMFG, a novel multi-scale interactive fusion graph neural network for spatial transcriptomics data clustering to identify spatial domains in tissues. Its multi-scale interactive fusion mechanism deeply integrates spatial and gene expression information across network layers. With spatial constraints and contrastive learning, stMFG captures discriminative embeddings while preserving spatial continuity. Evaluations on DLPFC and breast cancer datasets show that stMFG outperforms state-of-the-art benchmarks, achieving gains of up to 14\% in ARI on complex tissues. Future work will focus on improving computational efficiency and integrating additional data modalities.

\bibliography{References}


\begin{thebibliography}{79}
\ifx \bisbn   \undefined \def \bisbn  #1{ISBN #1}\fi
\ifx \binits  \undefined \def \binits#1{#1}\fi
\ifx \bauthor  \undefined \def \bauthor#1{#1}\fi
\ifx \batitle  \undefined \def \batitle#1{#1}\fi
\ifx \bjtitle  \undefined \def \bjtitle#1{#1}\fi
\ifx \bvolume  \undefined \def \bvolume#1{\textbf{#1}}\fi
\ifx \byear  \undefined \def \byear#1{#1}\fi
\ifx \bissue  \undefined \def \bissue#1{#1}\fi
\ifx \bfpage  \undefined \def \bfpage#1{#1}\fi
\ifx \blpage  \undefined \def \blpage #1{#1}\fi
\ifx \burl  \undefined \def \burl#1{\textsf{#1}}\fi
\ifx \doiurl  \undefined \def \doiurl#1{\url{https://doi.org/#1}}\fi
\ifx \betal  \undefined \def \betal{\textit{et al.}}\fi
\ifx \binstitute  \undefined \def \binstitute#1{#1}\fi
\ifx \binstitutionaled  \undefined \def \binstitutionaled#1{#1}\fi
\ifx \bctitle  \undefined \def \bctitle#1{#1}\fi
\ifx \beditor  \undefined \def \beditor#1{#1}\fi
\ifx \bpublisher  \undefined \def \bpublisher#1{#1}\fi
\ifx \bbtitle  \undefined \def \bbtitle#1{#1}\fi
\ifx \bedition  \undefined \def \bedition#1{#1}\fi
\ifx \bseriesno  \undefined \def \bseriesno#1{#1}\fi
\ifx \blocation  \undefined \def \blocation#1{#1}\fi
\ifx \bsertitle  \undefined \def \bsertitle#1{#1}\fi
\ifx \bsnm \undefined \def \bsnm#1{#1}\fi
\ifx \bsuffix \undefined \def \bsuffix#1{#1}\fi
\ifx \bparticle \undefined \def \bparticle#1{#1}\fi
\ifx \barticle \undefined \def \barticle#1{#1}\fi
\bibcommenthead
\ifx \bconfdate \undefined \def \bconfdate #1{#1}\fi
\ifx \botherref \undefined \def \botherref #1{#1}\fi
\ifx \url \undefined \def \url#1{\textsf{#1}}\fi
\ifx \bchapter \undefined \def \bchapter#1{#1}\fi
\ifx \bbook \undefined \def \bbook#1{#1}\fi
\ifx \bcomment \undefined \def \bcomment#1{#1}\fi
\ifx \oauthor \undefined \def \oauthor#1{#1}\fi
\ifx \citeauthoryear \undefined \def \citeauthoryear#1{#1}\fi
\ifx \endbibitem  \undefined \def \endbibitem {}\fi
\ifx \bconflocation  \undefined \def \bconflocation#1{#1}\fi
\ifx \arxivurl  \undefined \def \arxivurl#1{\textsf{#1}}\fi
\csname PreBibitemsHook\endcsname

%%% 1
\bibitem[\protect\citeauthoryear{Zormpas et~al.}{2023}]{b1}
\begin{barticle}
\bauthor{\bsnm{Zormpas}, \binits{E.}},
\bauthor{\bsnm{Queen}, \binits{R.}},
\bauthor{\bsnm{Comber}, \binits{A.}},
\bauthor{\bsnm{Cockell}, \binits{S.J.}}:
\batitle{Mapping the transcriptome: Realizing the full potential of spatial
  data analysis}.
\bjtitle{Cell}
\bvolume{186}(\bissue{26}),
\bfpage{5677}--\blpage{5689}
(\byear{2023})
\end{barticle}
\endbibitem

%%% 2
\bibitem[\protect\citeauthoryear{Xia et~al.}{2022}]{b2}
\begin{barticle}
\bauthor{\bsnm{Xia}, \binits{W.}},
\bauthor{\bsnm{Gao}, \binits{Q.}},
\bauthor{\bsnm{Wang}, \binits{Q.}},
\bauthor{\bsnm{Gao}, \binits{X.}},
\bauthor{\bsnm{Ding}, \binits{C.}},
\bauthor{\bsnm{Tao}, \binits{D.}}:
\batitle{Tensorized bipartite graph learning for multi-view clustering}.
\bjtitle{IEEE Transactions on Pattern Analysis and Machine Intelligence}
\bvolume{45}(\bissue{4}),
\bfpage{5187}--\blpage{5202}
(\byear{2022})
\end{barticle}
\endbibitem

%%% 3
\bibitem[\protect\citeauthoryear{Ismkhan}{2018}]{b3}
\begin{barticle}
\bauthor{\bsnm{Ismkhan}, \binits{H.}}:
\batitle{Ik-means-+: An iterative clustering algorithm based on an enhanced
  version of the k-means}.
\bjtitle{Pattern Recognition}
\bvolume{79},
\bfpage{402}--\blpage{413}
(\byear{2018})
\end{barticle}
\endbibitem

%%% 4
\bibitem[\protect\citeauthoryear{Yuan et~al.}{2024}]{b4}
\begin{botherref}
\oauthor{\bsnm{Yuan}, \binits{Y.}},
\oauthor{\bsnm{Wang}, \binits{Y.}},
\oauthor{\bsnm{Luo}, \binits{X.}}:
A node-collaboration-informed graph convolutional network for highly accurate
  representation to undirected weighted graph.
IEEE Transactions on Neural Networks and Learning Systems
(2024)
\end{botherref}
\endbibitem

%%% 5
\bibitem[\protect\citeauthoryear{Chen et~al.}{2024}]{b5}
\begin{barticle}
\bauthor{\bsnm{Chen}, \binits{J.}},
\bauthor{\bsnm{Yuan}, \binits{Y.}},
\bauthor{\bsnm{Luo}, \binits{X.}}:
\batitle{Sdgnn: Symmetry-preserving dual-stream graph neural networks}.
\bjtitle{IEEE/CAA journal of automatica sinica}
\bvolume{11}(\bissue{7}),
\bfpage{1717}--\blpage{1719}
(\byear{2024})
\end{barticle}
\endbibitem

%%% 6
\bibitem[\protect\citeauthoryear{Wang et~al.}{2024}]{b6}
\begin{botherref}
\oauthor{\bsnm{Wang}, \binits{L.}},
\oauthor{\bsnm{Liu}, \binits{K.}},
\oauthor{\bsnm{Yuan}, \binits{Y.}}:
Gt-a 2 t: Graph tensor alliance attention network.
IEEE/CAA Journal of Automatica Sinica
(2024)
\end{botherref}
\endbibitem

%%% 7
\bibitem[\protect\citeauthoryear{Han et~al.}{2025}]{b7}
\begin{bchapter}
\bauthor{\bsnm{Han}, \binits{M.}},
\bauthor{\bsnm{Wang}, \binits{L.}},
\bauthor{\bsnm{Yuan}, \binits{Y.}},
\bauthor{\bsnm{Luo}, \binits{X.}}:
\bctitle{Sgd-dyg: Self-reliant global dependency apprehending on dynamic
  graphs}.
In: \bbtitle{Proceedings of the 31st ACM SIGKDD Conference on Knowledge
  Discovery and Data Mining V. 2},
pp. \bfpage{802}--\blpage{813}
(\byear{2025})
\end{bchapter}
\endbibitem

%%% 8
\bibitem[\protect\citeauthoryear{Luo et~al.}{2023}]{b8}
\begin{barticle}
\bauthor{\bsnm{Luo}, \binits{X.}},
\bauthor{\bsnm{Wang}, \binits{L.}},
\bauthor{\bsnm{Hu}, \binits{P.}},
\bauthor{\bsnm{Hu}, \binits{L.}}:
\batitle{Predicting protein-protein interactions using sequence and network
  information via variational graph autoencoder}.
\bjtitle{IEEE/ACM Transactions on Computational Biology and Bioinformatics}
\bvolume{20}(\bissue{5}),
\bfpage{3182}--\blpage{3194}
(\byear{2023})
\end{barticle}
\endbibitem

%%% 9
\bibitem[\protect\citeauthoryear{Bi et~al.}{2023}]{b9}
\begin{barticle}
\bauthor{\bsnm{Bi}, \binits{F.}},
\bauthor{\bsnm{He}, \binits{T.}},
\bauthor{\bsnm{Xie}, \binits{Y.}},
\bauthor{\bsnm{Luo}, \binits{X.}}:
\batitle{Two-stream graph convolutional network-incorporated latent feature
  analysis}.
\bjtitle{IEEE Transactions on Services Computing}
\bvolume{16}(\bissue{4}),
\bfpage{3027}--\blpage{3042}
(\byear{2023})
\end{barticle}
\endbibitem

%%% 10
\bibitem[\protect\citeauthoryear{Xu et~al.}{2024}]{b10}
\begin{barticle}
\bauthor{\bsnm{Xu}, \binits{H.}},
\bauthor{\bsnm{Fu}, \binits{H.}},
\bauthor{\bsnm{Long}, \binits{Y.}},
\bauthor{\bsnm{Ang}, \binits{K.S.}},
\bauthor{\bsnm{Sethi}, \binits{R.}},
\bauthor{\bsnm{Chong}, \binits{K.}},
\bauthor{\bsnm{Li}, \binits{M.}},
\bauthor{\bsnm{Uddamvathanak}, \binits{R.}},
\bauthor{\bsnm{Lee}, \binits{H.K.}},
\bauthor{\bsnm{Ling}, \binits{J.}}, \betal:
\batitle{Unsupervised spatially embedded deep representation of spatial
  transcriptomics}.
\bjtitle{Genome Medicine}
\bvolume{16}(\bissue{1}),
\bfpage{12}
(\byear{2024})
\end{barticle}
\endbibitem

%%% 11
\bibitem[\protect\citeauthoryear{Wang et~al.}{2023}]{b11}
\begin{barticle}
\bauthor{\bsnm{Wang}, \binits{B.}},
\bauthor{\bsnm{Luo}, \binits{J.}},
\bauthor{\bsnm{Liu}, \binits{Y.}},
\bauthor{\bsnm{Shi}, \binits{W.}},
\bauthor{\bsnm{Xiong}, \binits{Z.}},
\bauthor{\bsnm{Shen}, \binits{C.}},
\bauthor{\bsnm{Long}, \binits{Y.}}:
\batitle{Spatial-mgcn: a novel multi-view graph convolutional network for
  identifying spatial domains with attention mechanism}.
\bjtitle{Briefings in Bioinformatics}
\bvolume{24}(\bissue{5}),
\bfpage{262}
(\byear{2023})
\end{barticle}
\endbibitem

%%% 12
\bibitem[\protect\citeauthoryear{Zhu et~al.}{2024}]{b12}
\begin{botherref}
\oauthor{\bsnm{Zhu}, \binits{Y.}},
\oauthor{\bsnm{He}, \binits{X.}},
\oauthor{\bsnm{Tang}, \binits{C.}},
\oauthor{\bsnm{Liu}, \binits{X.}},
\oauthor{\bsnm{Liu}, \binits{Y.}},
\oauthor{\bsnm{He}, \binits{K.}}:
Multi-view adaptive fusion network for spatially resolved transcriptomics data
  clustering.
IEEE Transactions on Knowledge and Data Engineering
(2024)
\end{botherref}
\endbibitem

%%% 13
\bibitem[\protect\citeauthoryear{Peng et~al.}{2021}]{b13}
\begin{bchapter}
\bauthor{\bsnm{Peng}, \binits{Z.}},
\bauthor{\bsnm{Liu}, \binits{H.}},
\bauthor{\bsnm{Jia}, \binits{Y.}},
\bauthor{\bsnm{Hou}, \binits{J.}}:
\bctitle{Attention-driven graph clustering network}.
In: \bbtitle{Proceedings of the 29th ACM International Conference on
  Multimedia},
pp. \bfpage{935}--\blpage{943}
(\byear{2021})
\end{bchapter}
\endbibitem

%%% 14
\bibitem[\protect\citeauthoryear{Chen et~al.}{2024}]{b14}
\begin{barticle}
\bauthor{\bsnm{Chen}, \binits{M.-S.}},
\bauthor{\bsnm{Zhu}, \binits{X.-R.}},
\bauthor{\bsnm{Lin}, \binits{J.-Q.}},
\bauthor{\bsnm{Wang}, \binits{C.-D.}}:
\batitle{Contrastive multiview attribute graph clustering with adaptive
  encoders}.
\bjtitle{IEEE Transactions on Neural Networks and Learning Systems}
\bvolume{36}(\bissue{4}),
\bfpage{7184}--\blpage{7195}
(\byear{2024})
\end{barticle}
\endbibitem

%%% 15
\bibitem[\protect\citeauthoryear{Yu et~al.}{2022}]{b15}
\begin{barticle}
\bauthor{\bsnm{Yu}, \binits{Z.}},
\bauthor{\bsnm{Lu}, \binits{Y.}},
\bauthor{\bsnm{Wang}, \binits{Y.}},
\bauthor{\bsnm{Tang}, \binits{F.}},
\bauthor{\bsnm{Wong}, \binits{K.C.}},
\bauthor{\bsnm{Li}, \binits{X.}}:
\batitle{Zinb-based graph embedding autoencoder for single-cell rna-seq
  interpretations}.
\bjtitle{Proceedings of the AAAI Conference on Artificial Intelligence}
\bvolume{36}(\bissue{4}),
\bfpage{4671}--\blpage{4679}
(\byear{2022})
\end{barticle}
\endbibitem

%%% 16
\bibitem[\protect\citeauthoryear{Yuan et~al.}{2025}]{b16}
\begin{barticle}
\bauthor{\bsnm{Yuan}, \binits{Y.}},
\bauthor{\bsnm{Lu}, \binits{S.}},
\bauthor{\bsnm{Luo}, \binits{X.}}:
\batitle{A proportional integral controller-enhanced non-negative latent factor
  analysis model}.
\bjtitle{IEEE/CAA Journal of Automatica Sinica}
\bvolume{12}(\bissue{6}),
\bfpage{1246}--\blpage{1259}
(\byear{2025})
\end{barticle}
\endbibitem

%%% 17
\bibitem[\protect\citeauthoryear{Yuan et~al.}{2024}]{b17}
\begin{barticle}
\bauthor{\bsnm{Yuan}, \binits{Y.}},
\bauthor{\bsnm{Luo}, \binits{X.}},
\bauthor{\bsnm{Zhou}, \binits{M.}}:
\batitle{Adaptive divergence-based non-negative latent factor analysis of
  high-dimensional and incomplete matrices from industrial applications}.
\bjtitle{IEEE Transactions on Emerging Topics in Computational Intelligence}
\bvolume{8}(\bissue{2}),
\bfpage{1209}--\blpage{1222}
(\byear{2024})
\end{barticle}
\endbibitem

%%% 18
\bibitem[\protect\citeauthoryear{Yuan et~al.}{2022}]{b18}
\begin{barticle}
\bauthor{\bsnm{Yuan}, \binits{Y.}},
\bauthor{\bsnm{Luo}, \binits{X.}},
\bauthor{\bsnm{Shang}, \binits{M.}},
\bauthor{\bsnm{Wang}, \binits{Z.}}:
\batitle{A kalman-filter-incorporated latent factor analysis model for
  temporally dynamic sparse data}.
\bjtitle{IEEE Transactions on Cybernetics}
\bvolume{53}(\bissue{9}),
\bfpage{5788}--\blpage{5801}
(\byear{2022})
\end{barticle}
\endbibitem

%%% 19
\bibitem[\protect\citeauthoryear{Yuan et~al.}{2023}]{b19}
\begin{barticle}
\bauthor{\bsnm{Yuan}, \binits{Y.}},
\bauthor{\bsnm{Wang}, \binits{R.}},
\bauthor{\bsnm{Yuan}, \binits{G.}},
\bauthor{\bsnm{Xin}, \binits{L.}}:
\batitle{An adaptive divergence-based non-negative latent factor model}.
\bjtitle{IEEE Transactions on Systems, Man, and Cybernetics: Systems}
\bvolume{53}(\bissue{10}),
\bfpage{6475}--\blpage{6487}
(\byear{2023})
\end{barticle}
\endbibitem

%%% 20
\bibitem[\protect\citeauthoryear{Yuan et~al.}{2020}]{b20}
\begin{barticle}
\bauthor{\bsnm{Yuan}, \binits{Y.}},
\bauthor{\bsnm{He}, \binits{Q.}},
\bauthor{\bsnm{Luo}, \binits{X.}},
\bauthor{\bsnm{Shang}, \binits{M.}}:
\batitle{A multilayered-and-randomized latent factor model for high-dimensional
  and sparse matrices}.
\bjtitle{IEEE transactions on big data}
\bvolume{8}(\bissue{3}),
\bfpage{784}--\blpage{794}
(\byear{2020})
\end{barticle}
\endbibitem

%%% 21
\bibitem[\protect\citeauthoryear{Xin et~al.}{2019}]{b21}
\begin{barticle}
\bauthor{\bsnm{Xin}, \binits{L.}},
\bauthor{\bsnm{Yuan}, \binits{Y.}},
\bauthor{\bsnm{Zhou}, \binits{M.}},
\bauthor{\bsnm{Liu}, \binits{Z.}},
\bauthor{\bsnm{Shang}, \binits{M.}}:
\batitle{Non-negative latent factor model based on $\beta$-divergence for
  recommender systems}.
\bjtitle{IEEE Transactions on Systems, Man, and Cybernetics: Systems}
\bvolume{51}(\bissue{8}),
\bfpage{4612}--\blpage{4623}
(\byear{2019})
\end{barticle}
\endbibitem

%%% 22
\bibitem[\protect\citeauthoryear{Yuan et~al.}{2020}]{b22}
\begin{bchapter}
\bauthor{\bsnm{Yuan}, \binits{Y.}},
\bauthor{\bsnm{Luo}, \binits{X.}},
\bauthor{\bsnm{Shang}, \binits{M.}},
\bauthor{\bsnm{Wu}, \binits{D.}}:
\bctitle{A generalized and fast-converging non-negative latent factor model for
  predicting user preferences in recommender systems}.
In: \bbtitle{Proceedings of The Web Conference 2020},
pp. \bfpage{498}--\blpage{507}
(\byear{2020})
\end{bchapter}
\endbibitem

%%% 23
\bibitem[\protect\citeauthoryear{Yuan et~al.}{2018}]{b23}
\begin{barticle}
\bauthor{\bsnm{Yuan}, \binits{Y.}},
\bauthor{\bsnm{Luo}, \binits{X.}},
\bauthor{\bsnm{Shang}, \binits{M.-S.}}:
\batitle{Effects of preprocessing and training biases in latent factor models
  for recommender systems}.
\bjtitle{Neurocomputing}
\bvolume{275},
\bfpage{2019}--\blpage{2030}
(\byear{2018})
\end{barticle}
\endbibitem

%%% 24
\bibitem[\protect\citeauthoryear{Chen et~al.}{}]{b24}
\begin{botherref}
\oauthor{\bsnm{Chen}, \binits{M.}},
\oauthor{\bsnm{Li}, \binits{T.}},
\oauthor{\bsnm{Lou}, \binits{J.}},
\oauthor{\bsnm{Luo}, \binits{X.}}:
Latent factorization of tensors incorporated battery cycle life prediction.',
  ieee.
CAA Journal of Automatica Sinica
\textbf{10}
\end{botherref}
\endbibitem

%%% 25
\bibitem[\protect\citeauthoryear{Wu et~al.}{2023}]{b25}
\begin{botherref}
\oauthor{\bsnm{Wu}, \binits{D.}},
\oauthor{\bsnm{Li}, \binits{Z.}},
\oauthor{\bsnm{Yu}, \binits{Z.}},
\oauthor{\bsnm{He}, \binits{Y.}},
\oauthor{\bsnm{Luo}, \binits{X.}}:
Robust low-rank latent feature analysis for spatiotemporal signal recovery.
IEEE Transactions on Neural Networks and Learning Systems
(2023)
\end{botherref}
\endbibitem

%%% 26
\bibitem[\protect\citeauthoryear{Wu et~al.}{2024}]{b26}
\begin{botherref}
\oauthor{\bsnm{Wu}, \binits{H.}},
\oauthor{\bsnm{Qiao}, \binits{Y.}},
\oauthor{\bsnm{Luo}, \binits{X.}}:
A fine-grained regularization scheme for nonnegative latent factorization of
  high-dimensional and incomplete tensors.
IEEE Transactions on Services Computing
(2024)
\end{botherref}
\endbibitem

%%% 27
\bibitem[\protect\citeauthoryear{Chen et~al.}{2024}]{b27}
\begin{barticle}
\bauthor{\bsnm{Chen}, \binits{J.}},
\bauthor{\bsnm{Liu}, \binits{K.}},
\bauthor{\bsnm{Luo}, \binits{X.}},
\bauthor{\bsnm{Yuan}, \binits{Y.}},
\bauthor{\bsnm{Sedraoui}, \binits{K.}},
\bauthor{\bsnm{Al-Turki}, \binits{Y.}},
\bauthor{\bsnm{Zhou}, \binits{M.}}:
\batitle{A state-migration particle swarm optimizer for adaptive latent factor
  analysis of high-dimensional and incomplete data}.
\bjtitle{IEEE/CAA Journal of Automatica Sinica}
\bvolume{11}(\bissue{11}),
\bfpage{2220}--\blpage{2235}
(\byear{2024})
\end{barticle}
\endbibitem

%%% 28
\bibitem[\protect\citeauthoryear{Zhong and Luo}{2021}]{b28}
\begin{bchapter}
\bauthor{\bsnm{Zhong}, \binits{Y.}},
\bauthor{\bsnm{Luo}, \binits{X.}}:
\bctitle{Alternating-direction-method of multipliers-based symmetric
  nonnegative latent factor analysis for large-scale undirected weighted
  networks}.
In: \bbtitle{2021 IEEE 17th International Conference on Automation Science and
  Engineering (CASE)},
pp. \bfpage{1527}--\blpage{1532}
(\byear{2021}).
\bcomment{IEEE}
\end{bchapter}
\endbibitem

%%% 29
\bibitem[\protect\citeauthoryear{Wu et~al.}{2023}]{b29}
\begin{barticle}
\bauthor{\bsnm{Wu}, \binits{D.}},
\bauthor{\bsnm{Zhang}, \binits{P.}},
\bauthor{\bsnm{He}, \binits{Y.}},
\bauthor{\bsnm{Luo}, \binits{X.}}:
\batitle{Mmlf: Multi-metric latent feature analysis for high-dimensional and
  incomplete data}.
\bjtitle{IEEE transactions on services computing}
\bvolume{17}(\bissue{2}),
\bfpage{575}--\blpage{588}
(\byear{2023})
\end{barticle}
\endbibitem

%%% 30
\bibitem[\protect\citeauthoryear{Li et~al.}{2023}]{b30}
\begin{botherref}
\oauthor{\bsnm{Li}, \binits{W.}},
\oauthor{\bsnm{Wang}, \binits{R.}},
\oauthor{\bsnm{Luo}, \binits{X.}}:
A generalized nesterov-accelerated second-order latent factor model for
  high-dimensional and incomplete data.
IEEE Transactions on Neural Networks and Learning Systems
(2023)
\end{botherref}
\endbibitem

%%% 31
\bibitem[\protect\citeauthoryear{Bi et~al.}{2023}]{b31}
\begin{barticle}
\bauthor{\bsnm{Bi}, \binits{F.}},
\bauthor{\bsnm{He}, \binits{T.}},
\bauthor{\bsnm{Luo}, \binits{X.}}:
\batitle{A fast nonnegative autoencoder-based approach to latent feature
  analysis on high-dimensional and incomplete data}.
\bjtitle{IEEE Transactions on Services Computing}
\bvolume{17}(\bissue{3}),
\bfpage{733}--\blpage{746}
(\byear{2023})
\end{barticle}
\endbibitem

%%% 32
\bibitem[\protect\citeauthoryear{Liu et~al.}{2023}]{b32}
\begin{barticle}
\bauthor{\bsnm{Liu}, \binits{Z.}},
\bauthor{\bsnm{Luo}, \binits{X.}},
\bauthor{\bsnm{Zhou}, \binits{M.}}:
\batitle{Symmetry and graph bi-regularized non-negative matrix factorization
  for precise community detection}.
\bjtitle{IEEE Transactions on Automation Science and Engineering}
\bvolume{21}(\bissue{2}),
\bfpage{1406}--\blpage{1420}
(\byear{2023})
\end{barticle}
\endbibitem

%%% 33
\bibitem[\protect\citeauthoryear{Wu et~al.}{2022}]{b33}
\begin{barticle}
\bauthor{\bsnm{Wu}, \binits{D.}},
\bauthor{\bsnm{Luo}, \binits{X.}},
\bauthor{\bsnm{He}, \binits{Y.}},
\bauthor{\bsnm{Zhou}, \binits{M.}}:
\batitle{A prediction-sampling-based multilayer-structured latent factor model
  for accurate representation to high-dimensional and sparse data}.
\bjtitle{IEEE transactions on neural networks and learning systems}
\bvolume{35}(\bissue{3}),
\bfpage{3845}--\blpage{3858}
(\byear{2022})
\end{barticle}
\endbibitem

%%% 34
\bibitem[\protect\citeauthoryear{Tang and Luo}{2025}]{b34}
\begin{barticle}
\bauthor{\bsnm{Tang}, \binits{P.}},
\bauthor{\bsnm{Luo}, \binits{X.}}:
\batitle{Neural tucker factorization}.
\bjtitle{IEEE/CAA Journal of Automatica Sinica}
\bvolume{12}(\bissue{2}),
\bfpage{475}--\blpage{477}
(\byear{2025})
\end{barticle}
\endbibitem

%%% 35
\bibitem[\protect\citeauthoryear{Wang et~al.}{2024}]{b35}
\begin{barticle}
\bauthor{\bsnm{Wang}, \binits{J.}},
\bauthor{\bsnm{Li}, \binits{W.}},
\bauthor{\bsnm{Luo}, \binits{X.}}:
\batitle{A distributed adaptive second-order latent factor analysis model}.
\bjtitle{IEEE/CAA Journal of Automatica Sinica}
\bvolume{11}(\bissue{11}),
\bfpage{2343}--\blpage{2345}
(\byear{2024})
\end{barticle}
\endbibitem

%%% 36
\bibitem[\protect\citeauthoryear{Luo et~al.}{2022}]{b36}
\begin{barticle}
\bauthor{\bsnm{Luo}, \binits{X.}},
\bauthor{\bsnm{Wu}, \binits{H.}},
\bauthor{\bsnm{Li}, \binits{Z.}}:
\batitle{Neulft: A novel approach to nonlinear canonical polyadic decomposition
  on high-dimensional incomplete tensors}.
\bjtitle{IEEE Transactions on Knowledge and Data Engineering}
\bvolume{35}(\bissue{6}),
\bfpage{6148}--\blpage{6166}
(\byear{2022})
\end{barticle}
\endbibitem

%%% 37
\bibitem[\protect\citeauthoryear{Luo et~al.}{2021a}]{b37}
\begin{barticle}
\bauthor{\bsnm{Luo}, \binits{X.}},
\bauthor{\bsnm{Zhong}, \binits{Y.}},
\bauthor{\bsnm{Wang}, \binits{Z.}},
\bauthor{\bsnm{Li}, \binits{M.}}:
\batitle{An alternating-direction-method of multipliers-incorporated approach
  to symmetric non-negative latent factor analysis}.
\bjtitle{IEEE Transactions on Neural Networks and Learning Systems}
\bvolume{34}(\bissue{8}),
\bfpage{4826}--\blpage{4840}
(\byear{2021})
\end{barticle}
\endbibitem

%%% 38
\bibitem[\protect\citeauthoryear{Luo et~al.}{2021b}]{b38}
\begin{barticle}
\bauthor{\bsnm{Luo}, \binits{X.}},
\bauthor{\bsnm{Zhou}, \binits{Y.}},
\bauthor{\bsnm{Liu}, \binits{Z.}},
\bauthor{\bsnm{Zhou}, \binits{M.}}:
\batitle{Fast and accurate non-negative latent factor analysis of
  high-dimensional and sparse matrices in recommender systems}.
\bjtitle{IEEE Transactions on Knowledge and Data Engineering}
\bvolume{35}(\bissue{4}),
\bfpage{3897}--\blpage{3911}
(\byear{2021})
\end{barticle}
\endbibitem

%%% 39
\bibitem[\protect\citeauthoryear{Wu et~al.}{2023}]{b39}
\begin{barticle}
\bauthor{\bsnm{Wu}, \binits{D.}},
\bauthor{\bsnm{He}, \binits{Y.}},
\bauthor{\bsnm{Luo}, \binits{X.}}:
\batitle{A graph-incorporated latent factor analysis model for high-dimensional
  and sparse data}.
\bjtitle{IEEE transactions on emerging topics in computing}
\bvolume{11}(\bissue{4}),
\bfpage{907}--\blpage{917}
(\byear{2023})
\end{barticle}
\endbibitem

%%% 40
\bibitem[\protect\citeauthoryear{Chen and Luo}{2023}]{b40}
\begin{barticle}
\bauthor{\bsnm{Chen}, \binits{L.}},
\bauthor{\bsnm{Luo}, \binits{X.}}:
\batitle{Tensor distribution regression based on the 3d conventional neural
  networks}.
\bjtitle{IEEE/CAA Journal of Automatica Sinica}
\bvolume{10}(\bissue{7}),
\bfpage{1628}--\blpage{1630}
(\byear{2023})
\end{barticle}
\endbibitem

%%% 41
\bibitem[\protect\citeauthoryear{Bi et~al.}{2023}]{b41}
\begin{barticle}
\bauthor{\bsnm{Bi}, \binits{F.}},
\bauthor{\bsnm{Luo}, \binits{X.}},
\bauthor{\bsnm{Shen}, \binits{B.}},
\bauthor{\bsnm{Dong}, \binits{H.}},
\bauthor{\bsnm{Wang}, \binits{Z.}}:
\batitle{Proximal alternating-direction-method-of-multipliers-incorporated
  nonnegative latent factor analysis}.
\bjtitle{IEEE/CAA Journal of Automatica Sinica}
\bvolume{10}(\bissue{6}),
\bfpage{1388}--\blpage{1406}
(\byear{2023})
\end{barticle}
\endbibitem

%%% 42
\bibitem[\protect\citeauthoryear{Liu et~al.}{2023}]{b42}
\begin{barticle}
\bauthor{\bsnm{Liu}, \binits{Z.}},
\bauthor{\bsnm{Yi}, \binits{Y.}},
\bauthor{\bsnm{Luo}, \binits{X.}}:
\batitle{A high-order proximity-incorporated nonnegative matrix
  factorization-based community detector}.
\bjtitle{IEEE Transactions on Emerging Topics in Computational Intelligence}
\bvolume{7}(\bissue{3}),
\bfpage{700}--\blpage{714}
(\byear{2023})
\end{barticle}
\endbibitem

%%% 43
\bibitem[\protect\citeauthoryear{Li et~al.}{2022}]{b43}
\begin{barticle}
\bauthor{\bsnm{Li}, \binits{W.}},
\bauthor{\bsnm{Luo}, \binits{X.}},
\bauthor{\bsnm{Yuan}, \binits{H.}},
\bauthor{\bsnm{Zhou}, \binits{M.}}:
\batitle{A momentum-accelerated hessian-vector-based latent factor analysis
  model}.
\bjtitle{IEEE Transactions on Services Computing}
\bvolume{16}(\bissue{2}),
\bfpage{830}--\blpage{844}
(\byear{2022})
\end{barticle}
\endbibitem

%%% 44
\bibitem[\protect\citeauthoryear{Mo et~al.}{2024}]{b44}
\begin{barticle}
\bauthor{\bsnm{Mo}, \binits{Y.}},
\bauthor{\bsnm{Li}, \binits{X.}},
\bauthor{\bsnm{Mei}, \binits{J.}}:
\batitle{Semi-supervised nonnegative matrix factorization with label
  propagation and constraint propagation}.
\bjtitle{Engineering Applications of Artificial Intelligence}
\bvolume{133},
\bfpage{108196}
(\byear{2024})
\end{barticle}
\endbibitem

%%% 45
\bibitem[\protect\citeauthoryear{Li et~al.}{2025}]{b45}
\begin{botherref}
\oauthor{\bsnm{Li}, \binits{X.}},
\oauthor{\bsnm{Mei}, \binits{J.}},
\oauthor{\bsnm{Mo}, \binits{Y.}}:
Semi-supervised structured nonnegative matrix factorization for anchor graph
  embedding.
Neurocomputing,
130222
(2025)
\end{botherref}
\endbibitem

%%% 46
\bibitem[\protect\citeauthoryear{Wu et~al.}{2022}]{b46}
\begin{barticle}
\bauthor{\bsnm{Wu}, \binits{D.}},
\bauthor{\bsnm{Zhang}, \binits{P.}},
\bauthor{\bsnm{He}, \binits{Y.}},
\bauthor{\bsnm{Luo}, \binits{X.}}:
\batitle{A double-space and double-norm ensembled latent factor model for
  highly accurate web service qos prediction}.
\bjtitle{IEEE Transactions on Services Computing}
\bvolume{16}(\bissue{2}),
\bfpage{802}--\blpage{814}
(\byear{2022})
\end{barticle}
\endbibitem

%%% 47
\bibitem[\protect\citeauthoryear{Li et~al.}{2022}]{b47}
\begin{barticle}
\bauthor{\bsnm{Li}, \binits{W.}},
\bauthor{\bsnm{Wang}, \binits{R.}},
\bauthor{\bsnm{Luo}, \binits{X.}},
\bauthor{\bsnm{Zhou}, \binits{M.}}:
\batitle{A second-order symmetric non-negative latent factor model for
  undirected weighted network representation}.
\bjtitle{IEEE Transactions on Network Science and Engineering}
\bvolume{10}(\bissue{2}),
\bfpage{606}--\blpage{618}
(\byear{2022})
\end{barticle}
\endbibitem

%%% 48
\bibitem[\protect\citeauthoryear{Chen et~al.}{2022}]{b48}
\begin{barticle}
\bauthor{\bsnm{Chen}, \binits{J.}},
\bauthor{\bsnm{Wang}, \binits{R.}},
\bauthor{\bsnm{Wu}, \binits{D.}},
\bauthor{\bsnm{Luo}, \binits{X.}}:
\batitle{A differential evolution-enhanced position-transitional approach to
  latent factor analysis}.
\bjtitle{IEEE Transactions on Emerging Topics in Computational Intelligence}
\bvolume{7}(\bissue{2}),
\bfpage{389}--\blpage{401}
(\byear{2022})
\end{barticle}
\endbibitem

%%% 49
\bibitem[\protect\citeauthoryear{Luo et~al.}{2021a}]{b49}
\begin{barticle}
\bauthor{\bsnm{Luo}, \binits{X.}},
\bauthor{\bsnm{Wu}, \binits{H.}},
\bauthor{\bsnm{Wang}, \binits{Z.}},
\bauthor{\bsnm{Wang}, \binits{J.}},
\bauthor{\bsnm{Meng}, \binits{D.}}:
\batitle{A novel approach to large-scale dynamically weighted directed network
  representation}.
\bjtitle{IEEE Transactions on Pattern Analysis and Machine Intelligence}
\bvolume{44}(\bissue{12}),
\bfpage{9756}--\blpage{9773}
(\byear{2021})
\end{barticle}
\endbibitem

%%% 50
\bibitem[\protect\citeauthoryear{Luo et~al.}{2021b}]{b50}
\begin{barticle}
\bauthor{\bsnm{Luo}, \binits{X.}},
\bauthor{\bsnm{Liu}, \binits{Z.}},
\bauthor{\bsnm{Jin}, \binits{L.}},
\bauthor{\bsnm{Zhou}, \binits{Y.}},
\bauthor{\bsnm{Zhou}, \binits{M.}}:
\batitle{Symmetric nonnegative matrix factorization-based community detection
  models and their convergence analysis}.
\bjtitle{IEEE Transactions on Neural Networks and Learning Systems}
\bvolume{33}(\bissue{3}),
\bfpage{1203}--\blpage{1215}
(\byear{2021})
\end{barticle}
\endbibitem

%%% 51
\bibitem[\protect\citeauthoryear{Song et~al.}{2022}]{b51}
\begin{barticle}
\bauthor{\bsnm{Song}, \binits{Y.}},
\bauthor{\bsnm{Li}, \binits{M.}},
\bauthor{\bsnm{Zhu}, \binits{Z.}},
\bauthor{\bsnm{Yang}, \binits{G.}},
\bauthor{\bsnm{Luo}, \binits{X.}}:
\batitle{Nonnegative latent factor analysis-incorporated and feature-weighted
  fuzzy double $ c $-means clustering for incomplete data}.
\bjtitle{IEEE Transactions on Fuzzy Systems}
\bvolume{30}(\bissue{10}),
\bfpage{4165}--\blpage{4176}
(\byear{2022})
\end{barticle}
\endbibitem

%%% 52
\bibitem[\protect\citeauthoryear{Wu et~al.}{2021a}]{b52}
\begin{barticle}
\bauthor{\bsnm{Wu}, \binits{D.}},
\bauthor{\bsnm{He}, \binits{Y.}},
\bauthor{\bsnm{Luo}, \binits{X.}},
\bauthor{\bsnm{Zhou}, \binits{M.}}:
\batitle{A latent factor analysis-based approach to online sparse streaming
  feature selection}.
\bjtitle{IEEE Transactions on Systems, Man, and Cybernetics: Systems}
\bvolume{52}(\bissue{11}),
\bfpage{6744}--\blpage{6758}
(\byear{2021})
\end{barticle}
\endbibitem

%%% 53
\bibitem[\protect\citeauthoryear{Wu et~al.}{2021b}]{b53}
\begin{barticle}
\bauthor{\bsnm{Wu}, \binits{D.}},
\bauthor{\bsnm{Shang}, \binits{M.}},
\bauthor{\bsnm{Luo}, \binits{X.}},
\bauthor{\bsnm{Wang}, \binits{Z.}}:
\batitle{An l 1-and-l 2-norm-oriented latent factor model for recommender
  systems}.
\bjtitle{IEEE Transactions on Neural Networks and Learning Systems}
\bvolume{33}(\bissue{10}),
\bfpage{5775}--\blpage{5788}
(\byear{2021})
\end{barticle}
\endbibitem

%%% 54
\bibitem[\protect\citeauthoryear{Luo et~al.}{2020}]{b54}
\begin{barticle}
\bauthor{\bsnm{Luo}, \binits{X.}},
\bauthor{\bsnm{Yuan}, \binits{Y.}},
\bauthor{\bsnm{Chen}, \binits{S.}},
\bauthor{\bsnm{Zeng}, \binits{N.}},
\bauthor{\bsnm{Wang}, \binits{Z.}}:
\batitle{Position-transitional particle swarm optimization-incorporated latent
  factor analysis}.
\bjtitle{IEEE Transactions on Knowledge and Data Engineering}
\bvolume{34}(\bissue{8}),
\bfpage{3958}--\blpage{3970}
(\byear{2020})
\end{barticle}
\endbibitem

%%% 55
\bibitem[\protect\citeauthoryear{Wolf et~al.}{2018}]{b55}
\begin{barticle}
\bauthor{\bsnm{Wolf}, \binits{F.A.}},
\bauthor{\bsnm{Angerer}, \binits{P.}},
\bauthor{\bsnm{Theis}, \binits{F.J.}}:
\batitle{Scanpy: large-scale single-cell gene expression data analysis}.
\bjtitle{Genome biology}
\bvolume{19}(\bissue{1}),
\bfpage{15}
(\byear{2018})
\end{barticle}
\endbibitem

%%% 56
\bibitem[\protect\citeauthoryear{Yuan et~al.}{2024}]{b56}
\begin{barticle}
\bauthor{\bsnm{Yuan}, \binits{Y.}},
\bauthor{\bsnm{Li}, \binits{J.}},
\bauthor{\bsnm{Luo}, \binits{X.}}:
\batitle{A fuzzy pid-incorporated stochastic gradient descent algorithm for
  fast and accurate latent factor analysis}.
\bjtitle{IEEE Transactions on Fuzzy Systems}
\bvolume{32}(\bissue{7}),
\bfpage{4049}--\blpage{4061}
(\byear{2024})
\end{barticle}
\endbibitem

%%% 57
\bibitem[\protect\citeauthoryear{Li et~al.}{2025}]{b57}
\begin{botherref}
\oauthor{\bsnm{Li}, \binits{J.}},
\oauthor{\bsnm{Yuan}, \binits{Y.}},
\oauthor{\bsnm{Luo}, \binits{X.}}:
Learning error refinement in stochastic gradient descent-based latent factor
  analysis via diversified pid controllers.
IEEE Transactions on Emerging Topics in Computational Intelligence
(2025)
\end{botherref}
\endbibitem

%%% 58
\bibitem[\protect\citeauthoryear{Liao et~al.}{2025}]{b58}
\begin{barticle}
\bauthor{\bsnm{Liao}, \binits{X.}},
\bauthor{\bsnm{Hoang}, \binits{K.}},
\bauthor{\bsnm{Luo}, \binits{X.}}:
\batitle{Local search-based anytime algorithms for continuous distributed
  constraint optimization problems}.
\bjtitle{IEEE/CAA Journal of Automatica Sinica}
\bvolume{12}(\bissue{1}),
\bfpage{288}--\blpage{290}
(\byear{2025})
\end{barticle}
\endbibitem

%%% 59
\bibitem[\protect\citeauthoryear{Qin et~al.}{2023}]{b59}
\begin{barticle}
\bauthor{\bsnm{Qin}, \binits{W.}},
\bauthor{\bsnm{Luo}, \binits{X.}},
\bauthor{\bsnm{Zhou}, \binits{M.}}:
\batitle{Adaptively-accelerated parallel stochastic gradient descent for
  high-dimensional and incomplete data representation learning}.
\bjtitle{IEEE Transactions on Big Data}
\bvolume{10}(\bissue{1}),
\bfpage{92}--\blpage{107}
(\byear{2023})
\end{barticle}
\endbibitem

%%% 60
\bibitem[\protect\citeauthoryear{Qin and Luo}{2023}]{b60}
\begin{barticle}
\bauthor{\bsnm{Qin}, \binits{W.}},
\bauthor{\bsnm{Luo}, \binits{X.}}:
\batitle{Asynchronous parallel fuzzy stochastic gradient descent for
  high-dimensional incomplete data representation}.
\bjtitle{IEEE Transactions on Fuzzy Systems}
\bvolume{32}(\bissue{2}),
\bfpage{445}--\blpage{459}
(\byear{2023})
\end{barticle}
\endbibitem

%%% 61
\bibitem[\protect\citeauthoryear{Wei et~al.}{2022}]{b61}
\begin{barticle}
\bauthor{\bsnm{Wei}, \binits{L.}},
\bauthor{\bsnm{Jin}, \binits{L.}},
\bauthor{\bsnm{Luo}, \binits{X.}}:
\batitle{A robust coevolutionary neural-based optimization algorithm for
  constrained nonconvex optimization}.
\bjtitle{IEEE Transactions on Neural Networks and Learning Systems}
\bvolume{35}(\bissue{6}),
\bfpage{7778}--\blpage{7791}
(\byear{2022})
\end{barticle}
\endbibitem

%%% 62
\bibitem[\protect\citeauthoryear{Hu et~al.}{2023}]{b62}
\begin{barticle}
\bauthor{\bsnm{Hu}, \binits{L.}},
\bauthor{\bsnm{Yang}, \binits{Y.}},
\bauthor{\bsnm{Tang}, \binits{Z.}},
\bauthor{\bsnm{He}, \binits{Y.}},
\bauthor{\bsnm{Luo}, \binits{X.}}:
\batitle{Fcan-mopso: An improved fuzzy-based graph clustering algorithm for
  complex networks with multiobjective particle swarm optimization}.
\bjtitle{IEEE Transactions on Fuzzy Systems}
\bvolume{31}(\bissue{10}),
\bfpage{3470}--\blpage{3484}
(\byear{2023})
\end{barticle}
\endbibitem

%%% 63
\bibitem[\protect\citeauthoryear{Li et~al.}{2025}]{b63}
\begin{botherref}
\oauthor{\bsnm{Li}, \binits{X.}},
\oauthor{\bsnm{Mo}, \binits{Y.}},
\oauthor{\bsnm{Jiang}, \binits{H.}}:
A new adaptive two-parameter conjugate gradient method.
Engineering Optimization,
1--15
(2025)
\end{botherref}
\endbibitem

%%% 64
\bibitem[\protect\citeauthoryear{Lv et~al.}{2024}]{b64}
\begin{barticle}
\bauthor{\bsnm{Lv}, \binits{Z.}},
\bauthor{\bsnm{Song}, \binits{Y.}},
\bauthor{\bsnm{He}, \binits{C.}},
\bauthor{\bsnm{Xu}, \binits{L.}}:
\batitle{Remaining useful life prediction for lithium-ion batteries
  incorporating spatio-temporal information}.
\bjtitle{Journal of Energy Storage}
\bvolume{88},
\bfpage{111626}
(\byear{2024})
\end{barticle}
\endbibitem

%%% 65
\bibitem[\protect\citeauthoryear{Lv et~al.}{2025}]{b65}
\begin{barticle}
\bauthor{\bsnm{Lv}, \binits{Z.}},
\bauthor{\bsnm{Song}, \binits{Y.}},
\bauthor{\bsnm{Xue}, \binits{Y.}},
\bauthor{\bsnm{Xu}, \binits{S.}},
\bauthor{\bsnm{He}, \binits{C.}},
\bauthor{\bsnm{Xu}, \binits{L.}}:
\batitle{State estimation of lithium-ion batteries with state space model}.
\bjtitle{Engineering Applications of Artificial Intelligence}
\bvolume{159},
\bfpage{111463}
(\byear{2025})
\end{barticle}
\endbibitem

%%% 66
\bibitem[\protect\citeauthoryear{Lv et~al.}{2023}]{b66}
\begin{bchapter}
\bauthor{\bsnm{Lv}, \binits{Z.}},
\bauthor{\bsnm{He}, \binits{C.}},
\bauthor{\bsnm{Xu}, \binits{L.}}:
\bctitle{A study of chinese medicine entity recognition method by fusing
  multi-features and pointer networks}.
In: \bbtitle{2023 IEEE International Conference on Systems, Man, and
  Cybernetics (SMC)},
pp. \bfpage{2087}--\blpage{2092}
(\byear{2023}).
\bcomment{IEEE}
\end{bchapter}
\endbibitem

%%% 67
\bibitem[\protect\citeauthoryear{Wang et~al.}{2024}]{b67}
\begin{botherref}
\oauthor{\bsnm{Wang}, \binits{H.}},
\oauthor{\bsnm{Song}, \binits{Y.I.}},
\oauthor{\bsnm{Chen}, \binits{W.}},
\oauthor{\bsnm{Luo}, \binits{Z.}},
\oauthor{\bsnm{Chongshou}, \binits{L.I.}},
\oauthor{\bsnm{Tianrui}, \binits{L.I.}}:
A survey of co-clustering.
ACM transactions on knowledge discovery from data
(9),
18
(2024)
\end{botherref}
\endbibitem

%%% 68
\bibitem[\protect\citeauthoryear{Zhang et~al.}{2022}]{b68}
\begin{barticle}
\bauthor{\bsnm{Zhang}, \binits{F.}},
\bauthor{\bsnm{Jin}, \binits{L.}},
\bauthor{\bsnm{Luo}, \binits{X.}}:
\batitle{Error-summation enhanced newton algorithm for model predictive control
  of redundant manipulators}.
\bjtitle{IEEE Transactions on Industrial Electronics}
\bvolume{70}(\bissue{3}),
\bfpage{2800}--\blpage{2811}
(\byear{2022})
\end{barticle}
\endbibitem

%%% 69
\bibitem[\protect\citeauthoryear{Luo et~al.}{2024}]{b69}
\begin{barticle}
\bauthor{\bsnm{Luo}, \binits{X.}},
\bauthor{\bsnm{Chen}, \binits{J.}},
\bauthor{\bsnm{Yuan}, \binits{Y.}},
\bauthor{\bsnm{Wang}, \binits{Z.}}:
\batitle{Pseudo gradient-adjusted particle swarm optimization for accurate
  adaptive latent factor analysis}.
\bjtitle{IEEE Transactions on Systems, Man, and Cybernetics: Systems}
\bvolume{54}(\bissue{4}),
\bfpage{2213}--\blpage{2226}
(\byear{2024})
\end{barticle}
\endbibitem

%%% 70
\bibitem[\protect\citeauthoryear{Li et~al.}{2023}]{b70}
\begin{barticle}
\bauthor{\bsnm{Li}, \binits{J.}},
\bauthor{\bsnm{Luo}, \binits{X.}},
\bauthor{\bsnm{Yuan}, \binits{Y.}},
\bauthor{\bsnm{Gao}, \binits{S.}}:
\batitle{A nonlinear pid-incorporated adaptive stochastic gradient descent
  algorithm for latent factor analysis}.
\bjtitle{IEEE Transactions on Automation Science and Engineering}
\bvolume{21}(\bissue{3}),
\bfpage{3742}--\blpage{3756}
(\byear{2023})
\end{barticle}
\endbibitem

%%% 71
\bibitem[\protect\citeauthoryear{Chen et~al.}{2024}]{b71}
\begin{barticle}
\bauthor{\bsnm{Chen}, \binits{J.}},
\bauthor{\bsnm{Liu}, \binits{K.}},
\bauthor{\bsnm{Luo}, \binits{X.}},
\bauthor{\bsnm{Yuan}, \binits{Y.}},
\bauthor{\bsnm{Sedraoui}, \binits{K.}},
\bauthor{\bsnm{Al-Turki}, \binits{Y.}},
\bauthor{\bsnm{Zhou}, \binits{M.}}:
\batitle{A state-migration particle swarm optimizer for adaptive latent factor
  analysis of high-dimensional and incomplete data}.
\bjtitle{IEEE/CAA Journal of Automatica Sinica}
\bvolume{11}(\bissue{11}),
\bfpage{2220}--\blpage{2235}
(\byear{2024})
\end{barticle}
\endbibitem

%%% 72
\bibitem[\protect\citeauthoryear{Mei et~al.}{2024}]{b72}
\begin{barticle}
\bauthor{\bsnm{Mei}, \binits{J.}},
\bauthor{\bsnm{Li}, \binits{X.}},
\bauthor{\bsnm{Mo}, \binits{Y.}}:
\batitle{Dual semi-supervised hypergraph regular multi-view nmf with anchor
  graph embedding}.
\bjtitle{Knowledge-Based Systems}
\bvolume{305},
\bfpage{112662}
(\byear{2024})
\end{barticle}
\endbibitem

%%% 73
\bibitem[\protect\citeauthoryear{Knops et~al.}{2006}]{b73}
\begin{barticle}
\bauthor{\bsnm{Knops}, \binits{Z.F.}},
\bauthor{\bsnm{Maintz}, \binits{J.A.}},
\bauthor{\bsnm{Viergever}, \binits{M.A.}},
\bauthor{\bsnm{Pluim}, \binits{J.P.}}:
\batitle{Normalized mutual information based registration using k-means
  clustering and shading correction}.
\bjtitle{Medical image analysis}
\bvolume{10}(\bissue{3}),
\bfpage{432}--\blpage{439}
(\byear{2006})
\end{barticle}
\endbibitem

%%% 74
\bibitem[\protect\citeauthoryear{Luong et~al.}{2022}]{b74}
\begin{barticle}
\bauthor{\bsnm{Luong}, \binits{K.}},
\bauthor{\bsnm{Nayak}, \binits{R.}},
\bauthor{\bsnm{Balasubramaniam}, \binits{T.}},
\bauthor{\bsnm{Bashar}, \binits{M.A.}}:
\batitle{Multi-layer manifold learning for deep non-negative matrix
  factorization-based multi-view clustering}.
\bjtitle{Pattern Recognition}
\bvolume{131},
\bfpage{108815}
(\byear{2022})
\end{barticle}
\endbibitem

%%% 75
\bibitem[\protect\citeauthoryear{Liu et~al.}{2023}]{b75}
\begin{barticle}
\bauthor{\bsnm{Liu}, \binits{X.}},
\bauthor{\bsnm{Ding}, \binits{S.}},
\bauthor{\bsnm{Xu}, \binits{X.}},
\bauthor{\bsnm{Wang}, \binits{L.}}:
\batitle{Deep manifold regularized semi-nonnegative matrix factorization for
  multi-view clustering}.
\bjtitle{Applied Soft Computing}
\bvolume{132},
\bfpage{109806}
(\byear{2023})
\end{barticle}
\endbibitem

%%% 76
\bibitem[\protect\citeauthoryear{Wang et~al.}{2015}]{b76}
\begin{barticle}
\bauthor{\bsnm{Wang}, \binits{D.}},
\bauthor{\bsnm{Gao}, \binits{X.}},
\bauthor{\bsnm{Wang}, \binits{X.}}:
\batitle{Semi-supervised nonnegative matrix factorization via constraint
  propagation}.
\bjtitle{IEEE transactions on cybernetics}
\bvolume{46}(\bissue{1}),
\bfpage{233}--\blpage{244}
(\byear{2015})
\end{barticle}
\endbibitem

%%% 77
\bibitem[\protect\citeauthoryear{Liu et~al.}{2011}]{b77}
\begin{barticle}
\bauthor{\bsnm{Liu}, \binits{H.}},
\bauthor{\bsnm{Wu}, \binits{Z.}},
\bauthor{\bsnm{Li}, \binits{X.}},
\bauthor{\bsnm{Cai}, \binits{D.}},
\bauthor{\bsnm{Huang}, \binits{T.S.}}:
\batitle{Constrained nonnegative matrix factorization for image
  representation}.
\bjtitle{IEEE Transactions on Pattern Analysis and Machine Intelligence}
\bvolume{34}(\bissue{7}),
\bfpage{1299}--\blpage{1311}
(\byear{2011})
\end{barticle}
\endbibitem

%%% 78
\bibitem[\protect\citeauthoryear{Cai et~al.}{2019}]{b78}
\begin{barticle}
\bauthor{\bsnm{Cai}, \binits{H.}},
\bauthor{\bsnm{Liu}, \binits{B.}},
\bauthor{\bsnm{Xiao}, \binits{Y.}},
\bauthor{\bsnm{Lin}, \binits{L.}}:
\batitle{Semi-supervised multi-view clustering based on constrained nonnegative
  matrix factorization}.
\bjtitle{Knowledge-Based Systems}
\bvolume{182},
\bfpage{104798}
(\byear{2019})
\end{barticle}
\endbibitem

%%% 79
\bibitem[\protect\citeauthoryear{Zhou et~al.}{2021}]{b79}
\begin{barticle}
\bauthor{\bsnm{Zhou}, \binits{H.}},
\bauthor{\bsnm{Yin}, \binits{H.}},
\bauthor{\bsnm{Li}, \binits{Y.}},
\bauthor{\bsnm{Chai}, \binits{Y.}}:
\batitle{Multiview clustering via exclusive non-negative subspace learning and
  constraint propagation}.
\bjtitle{Information Sciences}
\bvolume{552},
\bfpage{102}--\blpage{117}
(\byear{2021})
\end{barticle}
\endbibitem

\end{thebibliography}

\end{document}